\documentclass[a4paper]{article}

\usepackage[english]{babel}
\usepackage[utf8]{inputenc}
\usepackage[T1]{fontenc}

\usepackage{listings}

\usepackage{lmodern}
\usepackage[autostyle]{csquotes}

\usepackage[a4paper,top=3cm,bottom=2cm,left=3cm,right=3cm,marginparwidth=1.75cm]{geometry}

\usepackage{amsmath}
\usepackage{graphicx}
\usepackage[colorinlistoftodos]{todonotes}
\usepackage{authblk}
\usepackage{rotating}
\usepackage{csvsimple}
\usepackage{float}

\usepackage[colorlinks=true, allcolors=blue]{hyperref}
\usepackage{rotating}

\title{\huge Readying Medical Students for Medical AI: \\The Need to Embed AI Ethics Education}
\author[1*]{Thomas P. Quinn}
\author[2*]{Simon Coghlan}

\affil[1]{\footnotesize Applied Artificial Intelligence Institute, Deakin University, Geelong, Australia}
\affil[2]{Centre for AI and Digital Ethics, School of Computing and Information Systems, The University of Melbourne, Melbourne, Australia

* \textit{contacttomquinn@gmail.com, simon.coghlan@unimelb.edu.au}
}
\date{}

\Affilfont{\fontsize{4}{4}}

\begin{document}
\maketitle

\begin{abstract}

Medical students will almost inevitably encounter powerful medical AI systems early in their careers. Yet, contemporary medical education does not adequately equip students with the basic clinical proficiency in medical AI needed to use these tools safely and effectively. Education reform is urgently needed, but not easily implemented, largely due to an already jam-packed medical curricula. In this article, we propose an education reform framework as an effective and efficient solution, which we call the \textit{Embedded AI Ethics Education Framework}. Unlike other calls for education reform to accommodate AI teaching that are more radical in scope, our framework is modest and incremental. It leverages existing bioethics or medical ethics curricula to develop and deliver content on the ethical issues associated with medical AI, especially the harms of technology misuse, disuse, and abuse that affect the risk-benefit analyses at the heart of healthcare. In doing so, the framework provides a simple tool for going beyond the \textit{``What?''} and the \textit{``Why?''} of medical AI ethics education, to answer the \textit{``How?''}, giving universities, course directors, and/or professors a broad road-map for equipping their students with the necessary clinical proficiency in medical AI.

\end{abstract}

\maketitle

\section{Introduction}

Medical AI is here. As of 2020, at least 16 machine learning algorithms have been approved for marketing by the FDA \cite{nagendran_artificial_2020}. Outside of hospitals, direct-to-consumer medical AI devices are already available on app stores and used by patients worldwide \cite{vayena_machine_2018}. More medical AI is on its way, and will affect areas as diverse as imaging, pathology, surgery, internal medicine, oncology, drug development, genomics, service delivery, safety monitoring, health monitoring, and medical record keeping \cite{amisha2019overview}. Disclosures of equity funding show a steady increase in worldwide medical AI investments to over \$500 million USD per quarter in 2018 \cite{mou_artificial_2019}, while medical data conducive to machine learning accumulates at unprecedented levels. Recognition of the need to teach AI literacy to future practitioners is occurring in other professional domains, such as in teaching and education \cite{kandlhofer2016artificial}. Because it is almost inevitable that current medical students will encounter powerful medical AI systems early in their careers, medical schools too need to begin readying their students now. In this paper, we investigate how this may best be done. 

Artificial intelligence--by which we roughly mean the replication or replacement of human cognition and learning with machine processes--promises great benefit, but also carries great risk. AI can be \textit{misused} when it is deployed inappropriately (e.g., due to over-trust), or \textit{disused} when it ought to be used but is not (e.g., due to under-trust) \cite{jacovi2021formalizing}. AI can also be \textit{abused} when employers and managers implement it without due regard for practitioners' capabilities and performance \cite{parasuraman_humans_1997}. To avoid medical AI misuse, disuse, and abuse harmful to patients and society, future doctors should be equipped with \textit{clinical proficiency in medical AI}. By ``clinical proficiency'', we mean not mastery of AI, but rather a level of technical-medical knowledge sufficient to understand, identify, and clearly communicate to patients the implications, risks, and benefits of medical AI technology across a wide range of clinical applications. A few calls have recently been made that medical students need to be taught clinical proficiency to meet their professional obligations as practitioners in what is being called the ``Age of Artificial Intelligence'' \cite{wartman_medical_2018}. Such professional obligations, many of which are codified by regulatory bodies\footnote{For example, \url{https://www.medicalboard.gov.au/codes-guidelines-policies/code-of-conduct.aspx}.}, include practising medicine safely and effectively for the primary good of the patient. Health professionals must be empowered to use (or avoid) medical AI in conformity with these fundamental social and fiduciary obligations.

While literature on medical AI has grown exponentially in the last few years, there are relatively few scholarly papers about teaching medical AI proficiency to medical students. Fewer papers still examine teaching medical AI \textit{ethics} to undergraduates and graduates. In a notable exception to this rule, Katznelson and Gerke recently identified and sought to justify key ethical ideas that medical students need to learn, as contextualized through real-life case studies involving AI \cite{katznelson_need_2021}. 
However, their recommendations mainly address the \textit{``What?''} and the \textit{``Why?''} of AI medical ethics education, discussing what content should be taught and why it should be taught. By contrast, we aim to address the urgent question of \textit{``How?''} clinical proficiency and ethical competency in medical AI could be introduced into existing medical curricula in a way that is both \textit{effective} and \textit{efficient}. 

In answer to this question, we propose the \textit{Embedded AI Ethics Education Framework}. Under this framework, clinical proficiency in medical AI is taught by integrating AI ethics education into existing bioethics or medical ethics coursework. Unlike alternative recommendations, which call for radical changes to the curricula \cite{kolachalama_machine_2018,wartman_medical_2018}, our framework aims to be both feasible and sufficient by narrowing the scope of medical AI education reform to be:
\begin{enumerate}
    \item \textbf{Effective:} Changes to medical education should focus on ethical issues, especially the ethical harms of AI misuse, disuse, and abuse at the heart of risk-benefit analysis.
    \item \textbf{Efficient:} Changes to medical education should be able to be readily implemented without any major disruption to the existing curriculum.
\end{enumerate}
We present our article in 3 sections that explore these points further. First, we introduce the embedded AI ethics education framework, and the motivation behind it. Second, we summarize the known ethical implications of medical AI, and argue why an embedded AI ethics education would promote clinical proficiency in medical AI (i.e., is \textit{effective}). Third, we reflect on the ongoing challenges associated with medical education reform, and argue why an embedded AI ethics education would address these challenges (i.e., is \textit{efficient}). Although we concentrate on educating medical and surgical practitioners, much of what we discuss will have at least some relevance to other healthcare professions that may incorporate AI. This discussion gives universities, course directors, and/or professors a broad road-map for equipping their students with the necessary clinical proficiency in medical AI.

\section{The Embedded AI Ethics Education Framework}

What is needed to reformulate medical education to give medical students clinical proficiency in medical AI? Some far-reaching proposals have been made. For example, Wartman and Combs propose a ``radical, not incremental'' overhaul of medical education that emphasizes ``collaboration with and management of AI'', alongside ``a better understanding of probabilities'' \cite{wartman_reimagining_2019}. Meanwhile, Kolachalama and Garg claim that all medical students should be taught some programming skills alongside how to analyze biomedical datasets with a range of machine learning techniques, typically at the hands of a data scientist instructor \cite{kolachalama_machine_2018}. However, these proposals are highly problematic and it is unclear whether they can (or even should) be implemented in the near future. As Kolachalama and Garg themselves acknowledge, introducing medical AI to medical curricula is enormously difficult because of the lack of accreditation requirements regarding AI, the ever-growing calls from a myriad of domain areas for new content in already overpacked curricula \cite{coghlan2019nutrition}, and the lack of faculty with expertise in AI \cite{kolachalama_machine_2018}.

\begin{figure}[H]
\centering
\scalebox{1}{
\includegraphics[width=(\textwidth)]{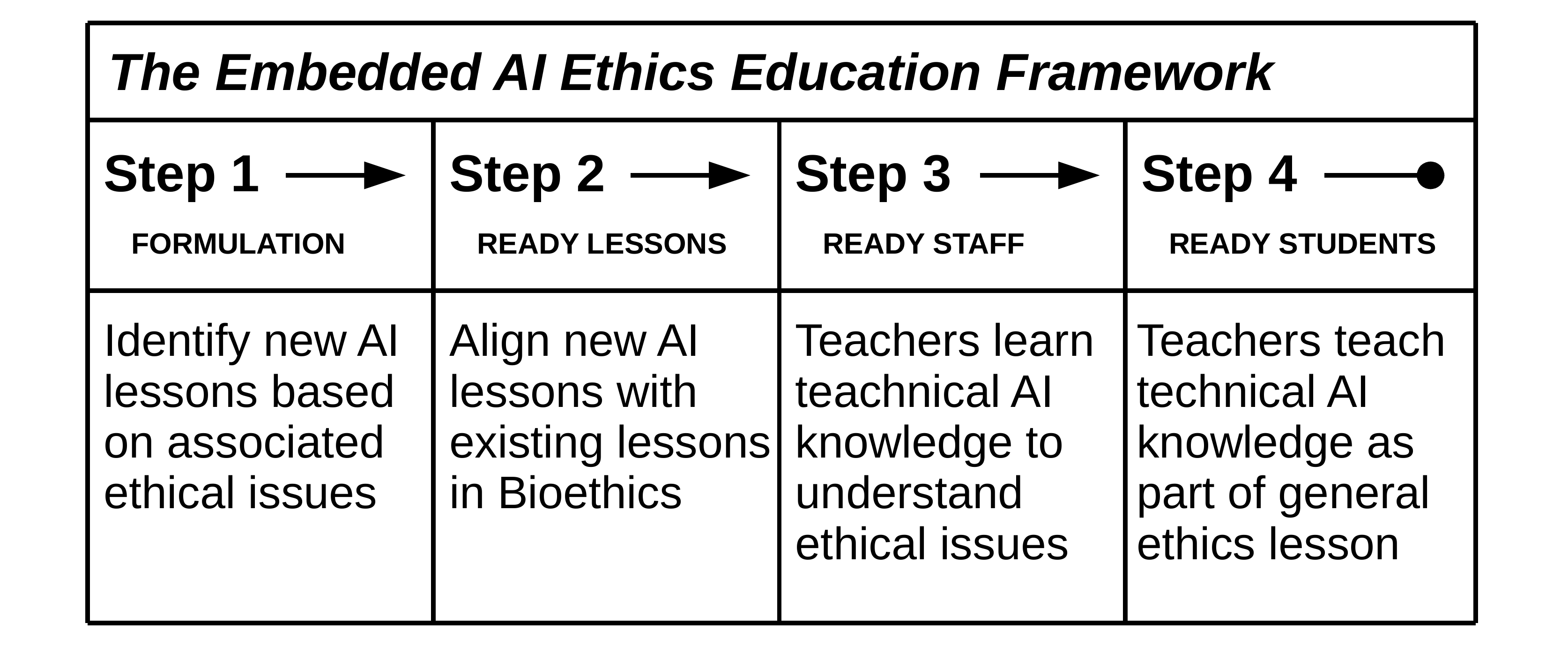}}
\caption{An outline of the Embedded AI Ethics Education Framework. We propose 4 steps as a simple tool for going beyond the \textit{``What?''} and the \textit{``Why?''} of medical AI ethics education, to answer the \textit{``How?''}, giving universities, course directors, and/or professors a broad road-map for equipping their students with clinical proficiency in medical AI.}
\label{fig:4steps}
\end{figure}

Yet education reform is urgent for the sake of patients and to earn society's trust in medical AI \cite{quinn_trust_2021}. To achieve this reform, medical students must understand certain technical-medical aspects of AI, including some elementary data science principles, so that they can confidently know when to use AI and how to explain its operation and outputs to their patients. Students also need to understand the ethical aspects of AI. Given the increasingly realistic possibility that AI will soon transform healthcare, both technical-medical and ethical competencies are essential. Other than doing nothing in medical education, this leaves several options available for medical schools to choose from:
\begin{enumerate}
    \item Teach technical-medical and ethical aspects \textit{separately}. Technical-medical content may be taught as a standalone subject, or distributed across a range of other subjects (e.g., endocrinology, cardio-pulmonary, geriatric care, etc.).
    \item Teach technical-medical and ethical aspects \textit{jointly} by integrating the ethical aspects into the technical-medical lessons.
    \item Teach technical-medical and ethical aspects \textit{jointly} by integrating the technical-medical aspects into the ethics lessons.
\end{enumerate}
Which of these is the most effective and efficient way to equip medical students with basic but sufficient clinical proficiency in medical AI? In the current circumstances, in which there is an urgent medical and ethical need to equip students, we favor Option 3. Option 3 does not require new technical-medical subjects but instead leverages existing educational frameworks. This less radical approach makes it far more feasible to introduce immediate changes to medical education. 
Our proposed \textit{Embedded AI Ethics Education Framework} implements Option 3 in 4 steps, as shown in Figure~\ref{fig:4steps}: (1) \textit{formulating} new AI lessons based on associated ethical issues; (2) \textit{readying lessons} by aligning the new AI ethics lessons with existing bioethics or medical ethics lessons; (3) \textit{readying staff} by enabling teachers to focus on learning the minimal prerequisite technical knowledge needed to fully understand the AI ethics lessons; and (4) \textit{readying students} by teaching them the same prerequisite technical knowledge as part of the AI ethics lesson.

In coming years, it may become necessary to give medical students more substantial technical training in AI, machine learning, and/or data science. 
Indeed, Katznelson and Gerke note that the AMA already recommends including data scientists and engineers in medical schools \cite{katznelson_need_2021}. However, we submit that the need to learn about medical AI 
in greater technical detail as part of the basic curriculum (i.e., from those with technical expertise in AI) will depend partly on the nature and extent of medical AI uptake in clinical practice. Moreover, as we have said, medical schools should do what is feasible and sufficient at the present time. Thus, we propose the Embedded AI Ethics Education Framework as a first step toward imparting clinical proficiency and ethical competency to professionals in medical AI, even if it is eventually superseded by future steps.

\section{The Effectiveness of Embedded AI Ethics Education}

According to the World Health Organization's first global report on artificial intelligence in health, AI should abide by the principles of (1) protecting autonomy, (2) promoting human well-being, safety, and public interest, (3) ensuring transparency, explainability, and intelligibility, (4) fostering responsibility and accountability, (5) ensuring inclusive and equitable care, and (6) promoting AI that is responsive and sustainable \cite{organization_ethics_2021}. AI ethics education presents a significant opportunity for future doctors to learn about how new technology and its use could either follow or depart from these principles. Although medical ethics teaching cannot hope to cover all of AI ethics (the literature on which is rapidly growing), it should at least cover some of it, and give future doctors the ability to reason critically and ethically about AI. Lessons could focus on any number of the ethical issues associated with medical AI, including (but not necessarily limited to): 
\begin{itemize}
    \item \textbf{informed consent:} It remains an open question as to whether there is something distinctive and essential about medical AI that requires informed consent about its use, and, if so, what aspects of medical AI need to be communicated as part of a sound informed consent process \cite{katznelson_need_2021,kiener_artificial_2020}.
    \item \textbf{bias:} Awareness is growing about the existence of sometimes obvious but also subtle biases within AI, and questions remain as to whether and how these biases can or should be mitigated \cite{katznelson_need_2021}.
    \item \textbf{safety:} New information technologies like AI can bring foreseen and unforeseen errors and risks to health delivery and care \cite{kim2017problems}. For example, medical AI devices are trained to minimize various mathematical ``loss functions'' (i.e., numerically defined objectives, such as average prediction error or classification error), and may not, through their explicit design and default use, necessarily minimize harm to patients \cite{katznelson_need_2021}.
    \item \textbf{transparency:} When medical AI is not understandable, physicians and patients may have to rely on model predictions that cannot be explicitly accounted for, which raises questions about which AI models should be used in which clinical contexts \cite{katznelson_need_2021}. 
    \item \textbf{patient privacy:} Since training medical AI models requires large amounts of clinical data, there are questions about what data should be shared and under what conditions. There are also concerns about the implications of collecting clinical data without patient awareness or informed consent \cite{katznelson_need_2021}.
    \item \textbf{allocation:} When using biased AI to allocate resources, it is possible that resources will be allocated in a discriminatory way. Alternatively, as medical AI becomes mainstream, it is possible that unequal access to beneficial but expensive medical AI could reinforce health hierarchies and inequities caused by wealth inequality \cite{katznelson_need_2021}.
    \item \textbf{quality assurance:} When using opaque ``black-box'' AI, errors and biases may not only exist, but go entirely undetected. Even when these errors do not impact patient safety, inexplicable errors may frustrate practitioners and patients, potentially undermining their trust in medical AI and leading to AI disuse \cite{quinn_trust_2021}.
    \item \textbf{trust:} Over-trust and under-trust in medical AI can both potentially have negative effects on patients, for example by causing AI misuse or disuse, respectively.\cite{jacovi2021formalizing}. Reflection on how this impacts clinic and hospital managers, professionals, and patients, and what can be done to enhance justified trust and distrust, may be required.
    \item \textbf{patient values:} Models may make recommendations that do not (or cannot) take into account patient preferences \cite{mcdougall2019computer}. For example, a model may recommend a chemotherapy option that prolongs quality of life over quantity (or \textit{vice versa}), implying a set of values that the patient may or may not hold. AI might also ``nudge'' patients and doctors in certain directions that may or may not be consistent with their values \cite{arnold2021teasing}.
    \item \textbf{cyber-security:} In addition to the cyber-security risks associated with computers more generally, AI presents altogether new cyber-security risks. For example, medical AI devices are susceptible to new kinds of cyber-attacks, called \textit{input attacks}, that could cause a model to make erroneous and potentially harmful predictions, even when the algorithm itself has not been compromised \cite{kiener_artificial_2020}.
    \item \textbf{confidentiality:} While patient privacy can be violated through various means, including cyberattacks or inadvertent leakage by hospital administrators, confidentiality itself is violated when practitioners themselves disclose patient information collected or stored digitally without patient consent (e.g., \cite{cohen_big_2019}).
    \item \textbf{justice:} If and as AI becomes more common and increasingly seen as indispensable in medicine, questions of fair access to its alleged benefits will become more acute. Potential injustices may arise for disadvantaged people who have less money, education, or access to healthcare \cite{katznelson_need_2021}.
    \item \textbf{human rights:} Access to AI might even be seen as part of a human right to healthcare \cite{jobin2019global}. Improving global health and wellbeing is one of the United Nations Sustainable Development Goals. However, some AI may become very expensive (e.g., due to the cost of the technology itself, as in the case of AI-guided robots, or due to the cost of administration, as in the case for a model requiring regular updates or audits), raising questions of access for patients in lower-income countries.
    \item \textbf{accountability:} Since medical AI carries risks and benefits for patient lives and wellbeing, there should presumably be systems in place to minimize risks and maximize benefits \cite{reddy2020governance}. What governance processes should there be to regulate it?
    \item \textbf{over-diagnosis and over-treatment:} Because AI can operate on masses of patient data (from medical records and potentially even sources like social media) to find new patterns and associations, it raises real risks of generating overdiagnoses \cite{vogt2019precision}, perhaps on a large scale. Overdiagnosis is the diagnosis of conditions that, if undiagnosed, would not have affected patient wellbeing. It can lead to over-treatment which can harm patients and waste healthcare resources.
    \item \textbf{automation bias:} Automation bias is a well-recognised cognitive phenomenon involving human over-reliance on automated processes \cite{goddard2012automation}. AI that makes seemingly rigorous autonomous judgments may entice users to succumb to a kind of of magical thinking about its epistemic authority. How might the medical profession deal with this risk?
    \item \textbf{skill erosion:} Reliance by doctors on AI might begin to erode important clinical skills, for instance, competencies in surgery, image reading, pathology, or diagnostic thinking \cite{arnold2021teasing}. To what degree and in what circumstances might replacing human thought and skill with machines be an ethical problem?
    \item \textbf{environmental sustainability of AI:} The medical profession is increasingly interested in the health implications of global warming, pollution, biodiversity loss, etc., and in the contribution of medicine itself to these enormous problems \cite{van2021sustainable}. Even now, many people do not appreciate the environmental impact that energy-intensive and mineral-hungry AI might have. A more expansive view of health, such as a One Health view \cite{coghlan2018one}, might contribute to a better understanding of AI, the environment, and wellbeing.
    \item \textbf{explainability/interpretability:} Deep neural networks (DNN) in particular can be inherently non-interpretable: the reasons underlying their decisions cannot in any direct way be rendered amenable to human understanding, let alone patient or doctor understanding. AI experts are working on ways of making DNN more explainable \cite{miller2017explainable}, but debate remains about the role of inherently opaque AI in medicine.
    \item \textbf{liability:} Medical AI can make recommendations ``autonomously.'' It can sometimes fail to give good recommendations due to: programming or other construction flaws, lack of proper documentation and user guidance, the manner in which healthcare professionals use it, the systems that are or are not in place in clinics or hospitals, and the state of hard and soft regulation in various jurisdictions \cite{bitterman2020approaching}. When medical AI harms patients, who is liable?
    \item \textbf{patient autonomy:} Respect for autonomy is a pillar of medical ethics. However, it is not entirely clear how patient autonomy might need to be supported with the advance of AI and whether we will need, say, a ``new concept of health-related digital autonomy'' \cite{laacke2020artificial}.
	\item \textbf{patient choice:} Some patients may wish not to use AI at this stage due to distrust, fear, etc.. To what extent should they be given choices and reminded of those choices, especially when high-stakes AI starts to becomes a more normal part of medical practice? Should we recognise a special right to refuse AI-based intervention? 
    \item \textbf{legal regulation:} Relatively new and sweeping laws, such as the EU's GDPR, are being introduced to limit the harms of AI \cite{floridi2019establishing}. However, it is often unclear exactly how novel laws, as well as existing ones in various jurisdictions, affect the use of medical AI. Moreover, ethics can both be affected by law and influence the development of new legal regulations.
    \item \textbf{information overload:} The overload of new information which health professionals already experience could be exacerbated with AI. Although AI aims to provide skill or decision support (or replacement) for doctors, it also has a learning curve and raises many new issues to navigate, as can be seen from this list. Information overload can affect the mental health of doctors, as well as their performance \cite{wartman_reimagining_2019}.
\end{itemize}
Although the notions of clinical proficiency and ethical competency are distinguishable, they are also profoundly connected. That is because, for a medical practitioner, ethical competency is often a precondition of clinical proficiency—and \textit{vice versa}.  Ethical issues in medicine concern, in significant part, properly identifying, weighing, and reasoning over the risks of harm and the promises of benefit in the light of the patient's good and their autonomous plans. A skilled practitioner who lacks ethical attitudes and reasoning ability can fall short of clinical proficiency by not serving the patient's good and even flouting professional guidelines. Conversely,  a well-intentioned practitioner who is clinically incompetent will not achieve ethical practice or meet their professional obligations.

\begin{figure}[h]
\centering
\scalebox{1}{
\includegraphics[width=(.8\textwidth)]{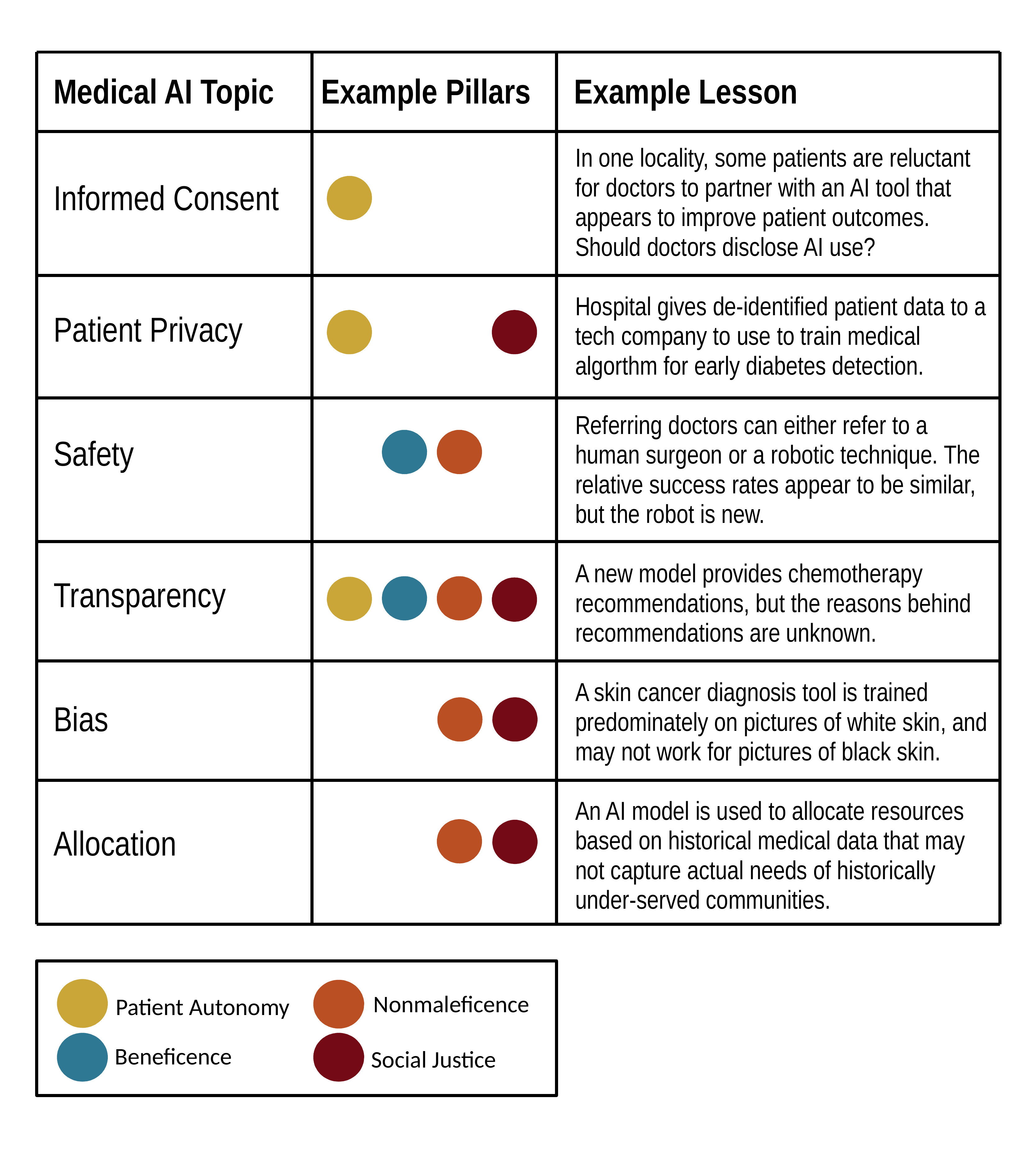}}
\caption{Six medical AI ethics topics identified by Katznelson and Gerke \cite{katznelson_need_2021}, along with their associated pillar and an example lesson. We present this table as an example of how new AI lessons can be identified by and aligned with existing lessons in bioethics, as suggested by Steps 1 and 2 of the Embedded AI Ethics Framework.}
\label{fig:4pillars}
\end{figure}

This logic applies to medical AI too. Future doctors should be able to identify potential consequences of AI and, when morally appropriate, adequately disclose them to, and discuss them with, their patients. Doctors should also be able to reason about the implications of AI in terms of established bioethical knowledge, such as Beauchamp and Childress's 4 pillars of medical ethics, namely nonmaleficence, respect for autonomy, beneficence, and justice \cite{beauchamp_principles_2001}. Thus, we propose the Embedded AI Ethics Education Framework as an effective tool for education reform because it addresses these imperatives directly. In Steps 1 and 2, new AI lessons are identified based on associated ethical issues, and then aligned with existing lessons in medical ethics (e.g., see Figure~\ref{fig:4pillars}). In Steps 3 and 4, staff first learn and then teach the minimal prerequisite technical knowledge needed to reason about the ethical implications of AI, giving medical students opportunities to acquire the ethical attitudes and reasoning ability needed to fulfill their clinical obligations.

\section{The Efficiency of Embedded AI Ethics Education}

If medical curricula are to be reformed, the changes must be feasible. Feasibility requires recognizing and overcoming key barriers to change. We group these barriers into 3 themes:
\begin{enumerate}
    \item \textbf{Barriers to curriculum reform:} Medical AI is a new field, and so it is not entirely clear what material should get taught; nor is it clear how best to teach it. This lack of clarity exists in a context where medical curricula are already jam-packed, making it difficult to add new content without commensurately removing existing content. For this reason, adding a whole new subject on medical AI may be intractable.
    \item \textbf{Barriers to curriculum delivery:} It may not be the case that any given university has staff with the relevant knowledge needed to teach the technical-medical aspects of medical AI. Given that it is not entirely clear what material should get taught, it is also not entirely clear what prerequisite knowledge staff should be expected to have.
    \item \textbf{Barriers to curriculum reception:} Depending on the chosen or required learning outcomes, students may require further prerequisite technical-medical knowledge, raising questions about how best to require, teach, and assess prerequisite knowledge. Notably, probability is a vast and complex subject \cite{wartman_medical_2018}, and it is not obvious how much probability medical students should have to learn before and/or during medical school.
\end{enumerate}
These barriers are large, but not insurmountable. By leveraging the existing medical ethics curriculum, universities can implement changes in a time-efficient and labour-efficient way. We propose the Embedded AI Education Framework as an efficient tool for education reform because it explicitly and directly addresses the identified barriers. The Steps of our framework each address a specific type of barrier.
\begin{enumerate}
    \item \textbf{Ready Lessons:} \textit{This Step identifies and aligns new AI lessons with existing lessons in bioethics or medical ethics, addressing barriers to curriculum reform.} For education reform to succeed, university staff must identify what new lessons are important for students to learn, then figure out how to fit these lessons into an already full curriculum. Focusing on the ethical issues associated with medical AI will (1) clarify how medical AI can harm and benefit patients; and (2) clarify what is important for students to know about the implications, risks, and benefits of medical AI technology. It is generally agreed upon that bioethical or medical ethics principles, such as those represented by the 4 so-called pillars, are worth upholding, so it is reasonable to use these pillars as a vehicle for medical AI education. Each time a pillar is mentioned, teachers have an opportunity to embed a medical AI lesson. This embedding does not require a revolutionary change to the existing curriculum. Instead, AI lessons are slotted into existing lessons without displacing swathes of important course material.
    \item \textbf{Ready Staff:} \textit{This Step enables teachers to focus on learning the minimal prerequisite technical knowledge needed to fully understand the AI ethics lesson(s), addressing barriers to curriculum delivery.}  Focusing on the ethical aspects of medical AI clarifies what prerequisite technical-medical knowledge staff need: staff only require the minimal technical-medical knowledge needed to fully understand the ethical issue. Consider the question, ``What do teachers need to know about opaque black-box AI?’’ According to our framework, the answer is, ``Enough to understand the ethical implications of its use’’. This sets a clear threshold by which universities can evaluate whether their staff possess adequate prerequisite knowledge to deliver medical AI lessons. The focus on ethics may also make it easier for staff to acquire missing technical-medical knowledge. Although ethics teachers may not have studied AI as a technical field, they have studied ethics, and thus may draw upon this knowledge to orient their further study and to retain the new knowledge.
    \item \textbf{Ready Students:} \textit{This Step teaches students more-or-less the same prerequisite technical knowledge needed to understand the AI ethics lesson(s) as part of the lesson itself, addressing barriers to curriculum reception.} Focusing on the ethical aspects of medical AI clarifies what prerequisite technical-medical knowledge students need: students, like staff, only require the minimal technical-medical knowledge needed to sufficiently understand the relevant ethical issues. As it does with staff, this approach sets a clear threshold that universities can use to evaluate whether their students have enough prerequisite knowledge. Notably, the level of technical detail required to understand the ethical issues, and to be able to successfully use and make decisions about AI in clinical practice, is relatively modest (though not negligible). The knowledge that teachers and students require is much less than the level of technical expertise that is needed for, say, designing and implementing medical AI. This threshold distinguishes our proposed framework from other proposals that have called for in-depth training in probability, programming, and machine learning. We do not think all medical students, at least in the short- to medium-term, need this type of training. Indeed, such extensive training is not only unnecessary for understanding the ethical issues and for being in a position to proficiently use AI in practice (which may require some on-the-job training for particular instances of AI even for those well-versed in relevant computer sciences), it is liable to cause information overload and displace vital parts of the medical curriculum.
\end{enumerate}

Although it appears unorthodox to teach about technical-medical concepts in medical \textit{ethics} programs, there are compelling reasons for doing so in this case. One issue for educators will be the \textit{novelty} of the field: AI ethics is a new discipline in a state of flux and without established ethical ideas and principles. Indeed, there are a number of ethical questions in this area about which there is no professional consensus, due in part to their novelty. For example, many medical AI applications rely on the use of opaque black-box models, and it is currently a wide open question as to whether black-box models should ever be used in high-stakes domains like medicine \cite{rudin_stop_2019}, or if so, under what conditions. Scholars and practitioners are only just starting to grapple with such questions, and only time will tell what the medical profession(s) chooses to do in its practice and policy. However, technological novelty, wide disagreement, and uncertainty are commonplace in ethical thinking and teaching and are often what make ethics important and interesting for students, teachers, and professionals. 

The current uncertainty and novelty of AI represents an excellent opportunity for inviting medical students to reason critically and ethically about healthcare. Indeed, medical students could be encouraged to engage with contemporary AI controversies and emerging regulator and professional standards, and to critically evaluate them. This could mean applying classic medical ethics principles, as well as moral theories and approaches like utilitarianism, ethics of care, virtue ethics, casuistry, and deontology to medical AI and to appropriate case studies. The very facts of novelty and unfamiliarity further highlights the significance of medical AI ethics education in its own right. In an age of autonomous machines that in some ways exceed human skill and thought, the medical profession's future will depend as much as ever on knowledgeable practitioners who can think critically about tough ethical questions created by medical innovation and technological change, and who can help forge a future where medical AI is used safely and effectively for the good of patients and society.

\section{Summary}

In this paper, we presented the Embedded AI Ethics Framework as an effective and efficient tool for preparing medical students for a future with medical AI . We then discussed the framework through 4 steps that address key barriers to education reform: (1) formulate new medical AI ethics lessons based on associated ethical issues, (2) ready lessons by aligning them with existing bioethics or medical ethics lessons, (3) ready staff by enabling them to learn the prerequisite technical knowledge needed to fully understand the new lessons, and (4) ready students by teaching them the same prerequisite technical knowledge as part of the ethics lesson. Unlike other proposals for education reform that call for widespread and radical change to medical education, our framework is more focused and incremental. It leverages existing bioethics curricula to teach medical students the technical-medical knowledge necessary and sufficient for them to understand, identify, and clearly communicate to future patients the implications, risks, and benefits of medical AI technology across a wide range of clinical and other applications. This renders the framework both feasible and sufficient for medical schools to implement now. Finally, the approach outlined above may be appropriated and adapted by other health disciplines also significantly affected by AI.

\bibliographystyle{unsrt}
\bibliography{references,simon}

\begin{thebibliography}{10}

\bibitem{nagendran_artificial_2020}
Myura Nagendran, Yang Chen, Christopher~A. Lovejoy, Anthony~C. Gordon, Matthieu
  Komorowski, Hugh Harvey, Eric~J. Topol, John P.~A. Ioannidis, Gary~S.
  Collins, and Mahiben Maruthappu.
\newblock Artificial intelligence versus clinicians: systematic review of
  design, reporting standards, and claims of deep learning studies.
\newblock {\em BMJ}, 368, March 2020.

\bibitem{vayena_machine_2018}
Effy Vayena, Alessandro Blasimme, and I.~Glenn Cohen.
\newblock Machine learning in medicine: {Addressing} ethical challenges.
\newblock {\em PLoS Medicine}, 15(11), November 2018.

\bibitem{amisha2019overview}
Paras~Malik Amisha, Monika Pathania, and Vyas~Kumar Rathaur.
\newblock Overview of artificial intelligence in medicine.
\newblock {\em Journal of family medicine and primary care}, 8(7):2328, 2019.

\bibitem{mou_artificial_2019}
Xiaomin Mou.
\newblock Artificial intelligence: {Investment} trends and selected industry
  uses.
\newblock Technical report, The World Bank, 2019.

\bibitem{kandlhofer2016artificial}
Martin Kandlhofer, Gerald Steinbauer, Sabine Hirschmugl-Gaisch, and Petra
  Huber.
\newblock Artificial intelligence and computer science in education: From
  kindergarten to university.
\newblock In {\em 2016 IEEE Frontiers in Education Conference (FIE)}, pages
  1--9. IEEE, 2016.

\bibitem{jacovi2021formalizing}
Alon Jacovi, Ana Marasovi{\'c}, Tim Miller, and Yoav Goldberg.
\newblock Formalizing trust in artificial intelligence: Prerequisites, causes
  and goals of human trust in ai.
\newblock In {\em Proceedings of the 2021 ACM Conference on Fairness,
  Accountability, and Transparency}, pages 624--635, 2021.

\bibitem{parasuraman_humans_1997}
Raja Parasuraman and Victor Riley.
\newblock Humans and {Automation}: {Use}, {Misuse}, {Disuse}, {Abuse}.
\newblock {\em Human Factors}, 39(2):230--253, June 1997.

\bibitem{wartman_medical_2018}
Steven~A. Wartman and C.~Donald Combs.
\newblock Medical {Education} {Must} {Move} {From} the {Information} {Age} to
  the {Age} of {Artificial} {Intelligence}.
\newblock {\em Academic Medicine: Journal of the Association of American
  Medical Colleges}, 93(8):1107--1109, August 2018.

\bibitem{katznelson_need_2021}
Gali Katznelson and Sara Gerke.
\newblock The need for health {AI} ethics in medical school education.
\newblock {\em Advances in Health Sciences Education}, March 2021.

\bibitem{kolachalama_machine_2018}
Vijaya~B. Kolachalama and Priya~S. Garg.
\newblock Machine learning and medical education.
\newblock {\em NPJ digital medicine}, 1:54, 2018.

\bibitem{wartman_reimagining_2019}
Steven~A. Wartman and C.~Donald Combs.
\newblock Reimagining {Medical} {Education} in the {Age} of {AI}.
\newblock {\em AMA journal of ethics}, 21(2):E146--152, February 2019.

\bibitem{coghlan2019nutrition}
Benjamin Coghlan, Simon Coghlan, and Alyce Wilson.
\newblock Nutrition education fit for modern health systems.
\newblock {\em The Lancet}, 394(10214):2071, 2019.

\bibitem{quinn_trust_2021}
Thomas~P Quinn, Manisha Senadeera, Stephan Jacobs, Simon Coghlan, and Vuong Le.
\newblock Trust and medical {AI}: the challenges we face and the expertise
  needed to overcome them.
\newblock {\em Journal of the American Medical Informatics Association},
  28(4):890--894, April 2021.

\bibitem{organization_ethics_2021}
World~Health Organization and {others}.
\newblock Ethics and governance of artificial intelligence for health: {WHO}
  guidance.
\newblock 2021.

\bibitem{kiener_artificial_2020}
Maximilian Kiener.
\newblock Artificial intelligence in medicine and the disclosure of risks.
\newblock {\em AI \& SOCIETY}, October 2020.

\bibitem{kim2017problems}
Mi~Ok Kim, Enrico Coiera, and Farah Magrabi.
\newblock Problems with health information technology and their effects on care
  delivery and patient outcomes: a systematic review.
\newblock {\em Journal of the American Medical Informatics Association},
  24(2):246--250, 2017.

\bibitem{mcdougall2019computer}
Rosalind~J McDougall.
\newblock Computer knows best? the need for value-flexibility in medical ai.
\newblock {\em Journal of medical ethics}, 45(3):156--160, 2019.

\bibitem{arnold2021teasing}
Mark~Henderson Arnold.
\newblock Teasing out artificial intelligence in medicine: An ethical critique
  of artificial intelligence and machine learning in medicine.
\newblock {\em Journal of Bioethical Inquiry}, 18(1):121--139, 2021.

\bibitem{cohen_big_2019}
I.~Glenn Cohen and Michelle~M. Mello.
\newblock Big {Data}, {Big} {Tech}, and {Protecting} {Patient} {Privacy}.
\newblock {\em JAMA}, 322(12):1141--1142, September 2019.

\bibitem{jobin2019global}
Anna Jobin, Marcello Ienca, and Effy Vayena.
\newblock The global landscape of ai ethics guidelines.
\newblock {\em Nature Machine Intelligence}, 1(9):389--399, 2019.

\bibitem{reddy2020governance}
Sandeep Reddy, Sonia Allan, Simon Coghlan, and Paul Cooper.
\newblock A governance model for the application of ai in health care.
\newblock {\em Journal of the American Medical Informatics Association},
  27(3):491--497, 2020.

\bibitem{vogt2019precision}
Henrik Vogt, Sara Green, Claus~Thorn Ekstr{\o}m, and John Brodersen.
\newblock How precision medicine and screening with big data could increase
  overdiagnosis.
\newblock {\em Bmj}, 366, 2019.

\bibitem{goddard2012automation}
Kate Goddard, Abdul Roudsari, and Jeremy~C Wyatt.
\newblock Automation bias: a systematic review of frequency, effect mediators,
  and mitigators.
\newblock {\em Journal of the American Medical Informatics Association},
  19(1):121--127, 2012.

\bibitem{van2021sustainable}
Aimee van Wynsberghe.
\newblock Sustainable ai: Ai for sustainability and the sustainability of ai.
\newblock {\em AI and Ethics}, pages 1--6, 2021.

\bibitem{coghlan2018one}
Simon Coghlan and Benjamin Coghlan.
\newblock One health, bioethics, and nonhuman ethics.
\newblock {\em The American Journal of Bioethics}, 18(11):3--5.

\bibitem{miller2017explainable}
Tim Miller, Piers Howe, and Liz Sonenberg.
\newblock Explainable ai: Beware of inmates running the asylum or: How i learnt
  to stop worrying and love the social and behavioural sciences.
\newblock {\em arXiv preprint arXiv:1712.00547}, 2017.

\bibitem{bitterman2020approaching}
Danielle~S Bitterman, Hugo~JWL Aerts, and Raymond~H Mak.
\newblock Approaching autonomy in medical artificial intelligence.
\newblock {\em The Lancet Digital Health}, 2(9):e447--e449, 2020.

\bibitem{laacke2020artificial}
Sebastian Laacke, Regina Mueller, Georg Schomerus, and Sabine Salloch.
\newblock Artificial intelligence, social media and depression. a new concept
  of health-related digital autonomy.
\newblock {\em The American Journal of Bioethics}, pages 1--33, 2020.

\bibitem{floridi2019establishing}
Luciano Floridi.
\newblock Establishing the rules for building trustworthy ai.
\newblock {\em Nature Machine Intelligence}, 1(6):261--262, 2019.

\bibitem{beauchamp_principles_2001}
Tom~L Beauchamp and James~F Childress.
\newblock {\em Principles of biomedical ethics}.
\newblock Oxford University Press, New York, N.Y., 2001.
\newblock OCLC: 758092388.

\bibitem{rudin_stop_2019}
Cynthia Rudin.
\newblock Stop explaining black box machine learning models for high stakes
  decisions and use interpretable models instead.
\newblock {\em Nature Machine Intelligence}, 1(5):206--215, May 2019.

\end{thebibliography}

\end{document}